\documentstyle[11pt,epsf,twocolumn]{article}
\begin{document}
\baselineskip=0.167in
\title{Free-ordered CUG on Chemical
Abstract Machine}
\author{{\Large Satoshi Tojo}\\
	 Mitsubishi Research Institute, Inc.\\
(e-mail) tojo@mri.co.jp}
\date{}
\maketitle

\begin{abstract}
We propose a paradigm for concurrent natural language generation. In
order to represent grammar rules distributively, we adopt categorial
unification grammar (CUG) where each category owns its functional
type. We augment typed lambda calculus with several new combinators,
to make the order of $\lambda$-conversions free for partial / local
processing. The concurrent calculus is modeled with {\it
Chemical Abstract Machine}.  We show an example of a Japanese {\em
causative} auxiliary verb that requires a drastic rearrangement of
case domination.
\end{abstract}

\section{Introduction}
Parallel and distributed computation is expected to be the main stream
of information processing.
In the conventional generation, the rules for composition are given
from the outside and those rules control all the behavior of the
symbols or the objects, for assembling a hierarchical tree
structure. For example, all the linguistic objects, such as words
and phrases must be applied to so-called grammar rules to form
grammatical structures or rational semantic representations, under a
strict controller process. However, this kind of formalization
obviously contradicts the partial / distributed processing that
would be required in parallel architecture in future.

In order to represent grammar rules
distributively, we adopt categorial grammar,
where we can an attach local grammar
rule to each word and phrase. What we aim in this paper is to
propose a paradigm that enables partial / local generation through
decompositions and reorganizations of tentative local structures.

In the following section, we introduce the extended
$\lambda$-calculus. Thereafter we introduce the ChAM model and we
reinterpret the model in terms of natural language processings.
Then we show the model of {\it membrane} interaction model with the
example of Japanese {\em causative} sentence that requires drastic
change of domination of cases. Finally we will discuss the future of
the model.

\section{Extended typed $\lambda$-calculus}
CUG (Categorial Unification Grammar)
\cite{uszkoreit86} is advantageous, compared to other phrase
structure grammars, for parallel architecture, because
we can regard categories as functional types and we can represent
grammar rules locally.
This means that we do not need
externally-given grammar rules but those rules reside within each word
or each phrase.
In this section, we
regard categories as polymorphic types and consider the
type calculus. In later sections we denote categories by DAG
(directed acyclic graph) of PATR grammar \cite{shieber86}.

\subsection{$\lambda$-calculus of polymorphic type}
We use greek letters, for type schemas. For type constants we use
$\sigma, \tau, \cdots$ while for type variables we use $\alpha, \beta,
\cdots$. $a:\alpha$ represents that the object $a$ is of type
$\alpha$. If $\alpha$ and $\beta$ are types, then $\alpha \rightarrow
\beta$ is a type.

The purpose of type inference is to infer the type of an object from a
set of objects whose types are known. We presuppose that two type
variables $\alpha$ and $\beta$ are unified with a unifier $\theta$. We
use $\Gamma$ for this set of type-known objects. The most important
two rules are as follows:
\footnote{
$\theta_{2}, \theta_{3}$ are for $\Gamma\theta_{2} \vdash t:\alpha
\rightarrow \beta$ and for $\Gamma\theta_{3} \vdash s:\alpha$,
respectively. $\theta_{4}$ unifies
$\alpha$ which appears in both type declarations.}
\begin{equation}
\frac{\Gamma\theta_{1} \cup \{x:\alpha\theta_{1} \} \vdash t:\beta}
{\Gamma\theta_{1} \vdash \lambda x^{\alpha}.t: \alpha\theta_{1} \rightarrow
\beta}\label{abs}\end{equation}
\begin{equation}
\frac{\Gamma\theta_{2}\theta_{3}\theta_{4} \vdash t:\alpha\theta_{4}
\rightarrow \beta\theta_{4} \hspace{9mm}
\Gamma\theta_{2}\theta_{3}\theta_{4} \vdash s:\alpha\theta_{4}}
{\Gamma\theta_{2}\theta_{3}\theta_{4} \vdash t(s):\beta\theta_{4}}
\label{app}\end{equation}
The rule (\ref{app}) corresponds to $\beta$-conversion of the ordinary
$\lambda$-calculus \cite{hindley}.

\subsection{Extended combinators}
In this subsection, we introduce two combinators that enable us to
change the order of $\lambda$-conversion, proposed by Steedman
\cite{steedman}, as a kind of type change \cite{Dowty88}.
The ordinary $\lambda$-calculus requires a strict order of
conversion. However, in a concurrent model, this kind of strict
order is a hindrance and contingent conversions are required.

{\bf C}-combinator changes the order of $\lambda$-variables as
follows:
\[{\bf C}(\lambda xy.f(x,y)) = \lambda yx.f(x,y).\]
Another requirement for exchanges of the order of $\lambda$-conversion
is the following case.
Suppose that we are required to compose all the following typed objects:
\[ \left\{ \begin{array}{l}
f : \beta \rightarrow \gamma \\
g : \alpha \rightarrow \beta \\
a : \alpha
\end{array} \right. \]
In such a case, we need to concatenate $g$ and $a$ first, and then $g(a)$
becomes applicable to $f$. However, with the help of the following {\bf
B}-combinator:
\[
{\bf B}(\lambda x.f(x))(\lambda y.g(y)) = \lambda x.f(g(x)).
\]
The $\lambda$-variable in $g$ can be shifted beyond the scope of $f$
so that we can concatenate $f$ and $g$ first, and, thus, have $a$ become
applicable as in Fig.~\ref{bf-C}.
\begin{figure}[htbp]
\begin{picture}(100,130)(-25,0)
\put(30,120){$\lambda y.g(y)$}\put(110,120){$a$}
\put(22,115){\line(1,0){100}}
\put(70,100){$g(a)$}\put(125,100){$\lambda x.f(x)$}
\put(65,95){\line(1,0){100}}
\put(100,80){$f(g(a))$}

\put(100,60){$\Downarrow$}

\put(30,40){$\lambda x.f(x)$}\put(100,40){$\lambda y.g(y)$}
\put(22,35){\line(1,0){120}} \put(10,32){${\bf B}$}
\put(70,20){$\lambda y.f(g(y))$}\put(150,20){$a$}
\put(65,15){\line(1,0){100}}
\put(100,0){$f(g(a))$}
\end{picture}
\caption{{\bf B}-combinator\label{bf-C}}
\end{figure}

\subsection{Cost of unification}
The repeated use of {\bf C}- and {\bf B}-combinators is still
problematic if we consider implementing it as an actual system
because the termination of processing is not guaranteed.  We have
modeled the process of a partial decomposition as an abstraction of
an argument of the first-order term. If this abstraction occurs
randomly, the process easily falls into a loop.  In order to avoid
this, we assume the unification cost.  If a compound term (a
subtree) were to be decomposed once, the element with the longer
distance should be abstracted first.  We can regard the whole
sentence structure as more grammatical if the sum of these
unification costs is smaller.  We introduce the heuristic costs
\cite{tojo91a}, considering the parallelism between syntactic cases
and semantic roles, as follows:
\[\begin{array}{ll}
\begin{array}{ll}
\| \theta_{(nom, agt)} \| = 1\\
\| \theta_{(dat, cgt)} \|  = 1 \\
\| \theta_{(acc, obj)}\| = 1
\end{array}&
\begin{array}{l}
\| \theta_{(obj, nom)} \| = k\\
\| \theta_{(dat, agt)} \| = k\\
\| \theta_{(t, obj)} \| = k
\end{array}\\
\\
\begin{array}{l}
\| \theta_{(nom, dat)} \| = \infty \\
\| \theta_{(agt, cgt)} \| = \infty \\
\hspace{15mm} \vdots
\end{array}&\end{array}\]
where $\theta_{(x,y)}$ represents a unifier of two DAG's: one's syntactic
case is $x$ and the other's semantic role is $y$.
$k$ is some constant larger than 1 ($k > 1$).

\section{Chemical Abstract Machine}
{\it Chemical Abstract
Machine} (ChAM, for short) \cite{berry90} is a paradigm of concurrent
$\lambda$-calculus. In this paper, we will
mention our principles on natural language processing with regard to
the ChAM model.

We assume the process of natural language recognition as follows.
Whenever a linguistic object is recognized, it is thrown into the
{\it solution} of ChAM, and acts as a {\it molecule}. Verbs and some
other auxiliary verbs introduces {\it membranes}. These membranes
becomes their scopes for case (or role) domination; namely, each verb
searches for molecules (noun phrases) that are necessary to satisfy
each verb's case (role) frame, within its membrane. In some
occasions, if multiple verbs exist in one sentence, they may conflict
as to which verb dominates which noun phrase. In such a
case, two membranes can interact and can exchange some molecules.

We use $ s_{1}, s_{2}, s_{3}, \cdots $
for membranes. When a membrane $s_i$ contains a molecule $\alpha$, we
denote as $ s_{i} \models \alpha.$
The supporting relation ($\models$) can be interpreted as an
inclusion relation ($\supset$) in this case. Two membranes can
interact when they contact with the notation `$\|$', as $ s_{1} \| s_{2}$.
If there is a floating molecule (that which is not yet concatenated
with other molecules) on one side, it can move through the porous
membranes. Valences for concatenation of each molecule are
represented by typed $lambda$-variables. If one membrane contains only one
composite structure, and it still has surplus valences, we can regard
that whole the membrane has those surplus valences as follows.
\[\begin{array}{c}
s_{2} \models \lambda xyz.make(x,y,z)\\
\downarrow \\
\lambda xyz. s_{2} \models make(x,y,z)
\end{array}\]

Now, we will apply our notions above to the actual problem of sentence
generation.

\section{Example: Japanese causative sentence}
In the Japanese
language, the causative and the change of voice are realized by
agglutinations of those auxiliary verbs at the tail of current verbs.
These auxiliary verbs as well as ordinary verbs can dominate some
cases so that these agglutinations may change the whole syntax \cite{gunji87}.
Namely the scope of the operation of these auxiliary verbs is
not the operated verb but the whole sentence.
In order to illustrate these role changes, we show the alternation of
the agent of the main verb in Table ~\ref{alter} with
a short tip to Japanese lexicon.
\begin{table*}[htbp]
\begin{center}
\begin{tabular}{l|l}
\hspace*{20mm}({\it yom-} $=$ read) & Who reads? \\ \hline
{\it Ken-wa Naomi-ni hon-wo yom-u.}&{\it Ken}\\
{\it Ken-wa Naomi-ni hon-wo yom-ase-ru.}&{\it Naomi}\\
{\it Ken-wa Naomi-ni hon-wo yom-are-ru.}&{\it Naomi}\\
{\it Ken-wa Naomi-ni hon-wo yom-ase-(r)-are-ru.}&{\it Ken}
\end{tabular}

\vspace*{2mm}

$\left(\begin{array}{ll}
\mbox{\it -wa}&\mbox{: nominative case marker}\\
\mbox{\it -ni}&\mbox{: dative case marker}\\
\mbox{\it -wo}&\mbox{: accusative case marker}\\
\mbox{\it hon}&\mbox{: noun for `book'}\\
\mbox{\it yom-}&\mbox{: root of verb `read'}\\
\mbox{\it -ase-}&\mbox{: auxiliary verb for causative}\\
\mbox{\it -are-}&\mbox{: auxiliary verb for passive}\\
\mbox{\it -(r)u}&\mbox{: present tense marker}
\end{array}\right)$
\end{center}
\caption{Agents
alternation by agglutination of auxiliary verbs\label{alter}}
\end{table*}

As an example, we will take the
sentence:
\begin{quote}
{\it Ken-wa Naomi-ni hon-wo yom-aseru.}\\
(Ken makes Naomi read the book.)
\end{quote}
First, we give DAG's for each lexical items in Fig~\ref{firstdags}.
\begin{figure*}[htbp]
\begin{center}
$\begin{array}{l}
W \models K (= Ken\mbox{-}wa) :
\left[\begin{array}{ll}cat&N\\case&nom\end{array}\right] \\
W \models N (= Naomi\mbox{-}ni) :
\left[\begin{array}{ll}cat&N\\case&dat\end{array}\right] \\
W \models B (= hon\mbox{-}wo) :
\left[\begin{array}{ll}cat&N\\case&acc\end{array}\right] \\
s_1 \models
\lambda xyz.read(x,y,z) (= yom\mbox{-}) :
\left[\begin{array}{ll}
val&\left[\begin{array}{ll}
cat & S\\
form & \left[\begin{array}{ll}
        form & fintie\end{array}\right]\\
\end{array}\right]\\
arg&\left[\begin{array}{ll}
cat & N\\
role & agent \end{array}\right]\\
arg&\left[\begin{array}{ll}
cat & N\\
role & object \end{array}\right]\\
arg&\left[\begin{array}{ll}
cat & N\\
role & co\mbox{-}agent\\
optionality & + \end{array}\right]
\end{array}\right] \end{array}$
\end{center}
\caption{Initial DAG}\label{firstdags}
\end{figure*}
The last DAG in Fig.~\ref{firstdags} represents that the verb `{\it
yomu} (read)' requires two roles `the reader' and `the object to be
read', and one optional role `the counter-agent' who hears what the
reader reads. In that figure, `$W \models$' means that each word is
recognized in the general {\it world} however a verb `{\it yomu}'
introduced a special membrane $s_1$ as a subworld of $W$. Each DAG
means a polymorphic type of the lexical item.

Assume that there is a parser that constructs partial tree structures,
as recognizing each word from the head sequentially. Then, when the
first four words are recognized, they can form a complete sentence of
(\ref{first-sentence}).
\begin{equation}
s_{1} \models \{read(K|_{\theta_1},N|_{\theta_2},B|_{\theta_3}):
\left[\begin{array}{ll}cat&S\end{array}\right]\}
\label{first-sentence}\end{equation}
Because all the three nouns are adequately concatenated by `read', a
sentential representation is made in the subworld of $s_1$. In
(\ref{first-sentence}), $\theta_i$'s are the records of unification,
that contain the costs and the original types; they become necessary when they
are backtracked, and in that meaning, those bindings are transitive.

Now, let us recapitulate what has occurred in the membrane $s_1$. There
were four lexical items in the set, and they are duly organized to a
sentence and $s_1$ becomes a singleton.
\[\begin{array}{rcl}
s_{1} &= &\{K:N,~ N:N, ~B:N,\\&& \lambda xyz.read(x,y,z):N\rightarrow N
\rightarrow N \rightarrow S\}\\
&&\downarrow \\
s_{1} &= &\{read(K,N,B)\}
\end{array}\]

Then, the problematic final word `{\it -aseru} (causative)' arrives;
its DAG representation is as in Fig.~\ref{problemdag}.
\begin{figure*}[htbp]
\begin{center}
$
\lambda xyz.make(x,y,z(y)):
\left[\begin{array}{ll}
val&\left[\begin{array}{ll}
cat & S\\
form & \left[\begin{array}{ll}
        form & fintie\end{array}\right]\\
\end{array}\right]\\
arg&\left[\begin{array}{ll}
cat & N\\
role & agent \end{array}\right]\\
arg&\left[\begin{array}{ll}
cat & N\\
role & co\mbox{-}Agent\\
optionality & + \end{array}\right]\\
arg&\left[\begin{array}{ll}
cat & S\\
subcat &\left[\begin{array}{ll}cat&N\\role&agent \end{array}\right]\\
role & t \end{array}\right]
\end{array}\right] $
\end{center}
\caption{DAG for `{\it make}'}\label{problemdag}
\end{figure*}
The DAG in Fig.~\ref{problemdag} requires a sentential form (category $S$) as
an argument, and in
addition, it subcategorizes an item of category $N$ as an agent of the
subsentence.

Now, the process becomes as in Fig.~\ref{process}.
\begin{figure*}[htbp]
\begin{center}
$\begin{array}{rcl}
s_{1} \models read(K|_{\theta_1},N|_{\theta_2},B|_{\theta_3}) &\| &
\lambda x'y'z'.s_{2} \models make(x',y',z') \\
&\downarrow&\\
\lambda x.s_{1} \models read(x, N|_{\theta_2}, B|_{\theta_3}) &\|&
\lambda y'z'. s_{2} \models make(K|_{\theta_4},y',z')\\
&\downarrow &\\
\lambda xy.s_{1} \models read(x,y,B) &\|&
\lambda z'. s_{2} \models make(K|_{\theta_4},N|_{\theta_5},z')\\
&\downarrow &\\
s_{1} &\|& s_{2}\models make(K|_{\theta_4},N|_{\theta_5},\lambda y.read(N,y,B))
\end{array}
$\end{center}
\caption{Process}\label{process}
\end{figure*}
All through the process in Fig.~\ref{process}, {\bf C}- and {\bf
B}-combinators are used repeatedly as well as ordinary type inference
({\ref{abs}) and (\ref{app}).  The second membrane $s_2$ requires an
{\it agent} role (the variable $x'$ of $make$). There is a record in
$\theta_1$ that it bit {\it agent}, so that the comparison should be
made between $\theta_1$ and $\theta_4 (= \theta_{(K, x')})$. However,
because both of $\theta_1$ and $\theta_4$ unifies nominative case and
agent role, the costs are equivalent.  In such a case, linguistic
heuristics will solve the problem. In this case, the agent of $make$
should be the nominative of the whole sentence, and the co-agent of
$make$ is the dative of the whole sentence, so that $K$ and $N$ are
bit by newly arrived $make$. $B$ remains bound to $read$, because
there is no $\lambda$-variable of that type in $make$.  The process is
depicted in fig.~\ref{eps2}.
\begin{figure*}[htbp]\begin{center}
\epsfxsize=13cm
\leavevmode\epsfbox{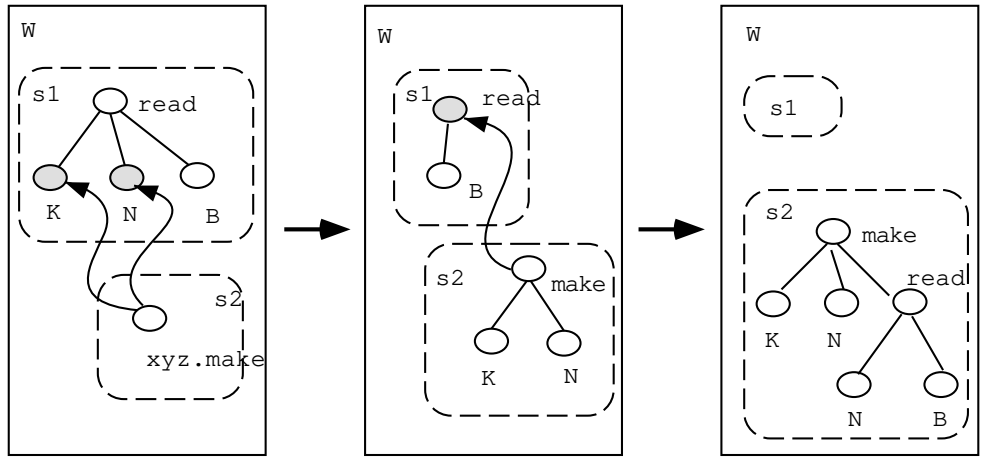}
\end{center}
\caption{Membranes interaction}
\label{eps2}
\end{figure*}

\section{Conclusion}
Introducing free-ordered typed $\lambda$-calculus, together with the
notion of unification costs in types, we have shown the structuring of
natural language syntax, by distributively represented types in random
orders.  We adopted a model of {\it Chemical Abstract Machine} for the
partial/ concurrent computation model.

Although we introduced the concept of costs and termination was
assured, the efficiency of constructing
a parsing tree would be far slower than sequential processing.
However our objective is not to propose a faster algorithm, but is to
show the possibility of distributed processing of natural languages.
We could show that natural language syntax is self-organizable, in
that each linguistic objects do not need to be poured into `molds',
viz., externally given grammar.

\bibliographystyle{plain}
\end{document}